# Deep Insights of Deepfake Technology : A Review

Bahar Uddin Mahmud*[1], Afsana Sharmin[2]

Dept. of CSE[1][2], Feni University[1],Feni, Bangladesh

Chittagong University of Engineering & Technology[2],Bangladesh

Email:mahmudbaharuddin@gmail.com*[1], afsana.cuet@gmail..com[2]

*Abstract*: Under the aegis ofof computer vision and deep learning technology, a new emerging techniques has introduced that anyone can make highly realistic but fake videos, images even can manipulates the voices. And this technology is widely known as Deepfake Technology. Although it seems interesting techniques to make fake videos or image of something or some individuals but it could spread as misinformation via internet. Deepfake contents could be dangerous for individuals as well as for our communities, organizations, countries religions etc. As Deepfake content creation involve a high level expertise with combination of several algorithms of deep learning, it seems almost real and genuine and difficult to differentiate. In this paper, a wide range of articles have been examined to understand Deepfake technology more extensively. We have examined several articles to find some insights such as what is Deepfake, who are the responsible for this, is there any benefits of Deepfake and what are the challenges of this technology. We have also examined several creation and detection techniques. Our results found that although Deepfake is threat to our societies, proper measures and strict regulations could prevent this.

## I. INTRODUCTION

DEEPFAKE, combination of deep learning and fake contents, is a process that involve swapping of a face from a person to a targeted person in a video and make the face expressing similar to targeted person and act like targeted person are saying those words that actually said by another person. Face swapping specially on image and video or manipulation of facial expression called as Deepfake methods [1]. Fake videos and images that go over the internet can easily exploit some individuals and it becomes a public issue recently [24]. Creating false contents by altering the face of an individual referring as source in an image or a video of another person referred as target is something that called DeepFake, which was named after a social media site Reddit account name "Deepfake" who then claimed to develop a machine learning technique to transfer celebrity faces into adult contents [5]. Furthermore, this technique is also used to create fake news, fraud and even spread hoaxes. Recently this field has got special concern of researcher who are now dedicatedly involve in breaking out the insights of Deepfake [6-8]. Deepfake contents can be pornography, political or bullying of a person by using his or her image and voice without his or her consent [9].There are several models in deep learning technology for instance, autoencoders and GAN to deal with several problems of computer vision domain [10-13].Several Deepfake algorithms have been proposed using generative adversarial networks to copy movements and facial expressions of a person and swap it with another person [14]. Political person, public figure, celebrity are the main victims of Deepfake. To spread false message of world leaders Deepfake technology used several times and it could be threat to world peace [15]. It can be used to mislead military personnel by providing fake image of maps and that could create serious damage to anyone [16]. Howeverr, there are many promising aspects of this technology. People who have lost their voice once can get it back with the help of this technology [17]. As we know false news spreads faster and wider but it is difficult task to recover it [18] . To know Deepfakes properly we have to know about it deeply such as what Deepfake actually is, how it comes, how to create it and how to detect it etc. As this field is almost new to researcher just introduced in 2017 there are not enough resources on this topics. Although several research have introduced recently to deal with social media misinformation related to Deepfake [19]. In this paper after the introduction ,we discussed more about Deepfake technology and its uses . And then we discussed possible threats and challenges of this technology, thereafter, we discussed different articles related to Deepfake generation and Deepfake detection and also its positive and negative sides . Finally, we discussed the limitations of our work, suggestions and future thoughts.

### A. Deepfake

Deepfake, a mixtures of deep learning and fake, are imitating contents where targeted subject's face was swapped by source person to make videos or images of target person [20-21]. Although, making of fake content is known to all but Deepfake is something beyond someone's thoughts which make this techniques more powerful and almost real using ML and AI to manipulate original content to make fraud one [22-24]. Deepfake has huge range of usages such as making fake pornography of well known celebrity, spreading of fake news, fake voices of politicians, financial fraud even many more [25-27]. Although face swapping technique is well known in film industry where several fake voice or videos made as their requirement but that takes huge time and certain level of expertise. But through deep learning techniques, anybody having sound computer knowledge and high configuration GPU computer can make trustworthy fake videos or images.

*B. Application of Deepfake*

Deepfake technology has a huge range of applications which could use both positively or negatively. Although most of the time it is used for malicious purposes. The unethical uses of Deepfake technology has harmful consequences in our society either in short term or long term. People regularly using social media are in a huge risk of Deepfake. However, proper use of this technology could bring many positive results. Below both negative and positive applications of Deepfake technology described in details.

*1) Negative Application:* Deepfake and technology related to this are expanding rapidly in current years. It has ample applications that use for malicious work against any human being specially against celebrity, political leaders. There are several reasons behind making Deepfake content that could be out of fun but sometimes it is used for taking revenge, blackmailing, stealing identity of someone and many more. There are thousands videos of Deepfake and most of them are adult videos of women without their permission [28]. Most common use of Deepfake technology is to make pornography of well known actress and it is rapidly increasing day by day specially of Hollywood actresses [29]. Moreover, in 2018 a software was built that make a women nude in a single click and it was widely went viral for malicious purposes to harass women [30]. The another most malicious use of Deepfake is to exploit world leaders and politician by making fake videos of them and sometimes it could have been great risk for world peace. Almost all world leader including Barack Obama, former president of USA, Donald trump, running president of USA, Nency Pelosi,USA based politician, Angela Merkel, German chancellor all are exploited by fake videos somehow and even Facebook founder Mark Zukerberg have faced similar occurance [31]. There are also vast use of Deepfake in Art, film industry and in social media.

*2) Positive Application:* Although most of the time this technology is used for malicious work with bad intention still it has some positive uses also in several sector. The Deepfake creation is no longer remain limited to experts, it is now become much more easier and accessible to anyone. Nowadays constructive uses of this technology widely increased. To create new art work, engage audiences and give them unique experiences this technology use a lot [32]. Recently the Dalí Museum in St. Petersburg, Florida created a chance to its visitor to meet Salvador Dalí and engage with his life more interactively to know this great personality via artificial intelligence[33]. Deepfake technology now use both as advertising and business purposes too. Technologists now are using the Deepfake to make copy of famous artwork such as creating video of famous monalisa artwork using the image [34]. Deepfake technology can save huge money and time of film industry by using the capabilities of Deepfake technology for editing videos rather than re-shot and there are many more positive examples such as famous footballer David Becham spoke in 9 different languages to run a campaign against malaria and it has also positive aspects in education sector [35].

GANS can be used in various field to give realistic experiences such as in retail sector, it might be possible to see the real product what we see in shop going physically [36]. Recently Reuters collaborated with AI startup Synthesis has made first ever synthesized news presenter by using Artificial intelligence techniques and it was done using same techniques that used in Deepfakes and it would be helpful for personalized news for individuals [37]. Deep generative models also has shown great possibilities of development in health care indus- try. To protect real data of patients and research work instead of sharing real data imaginary data could be generated via this technology [38]. Additionally, this technology has great potential in fundraising and awareness building by making videos of famous personality who is asking for help or fund for some novel work [39].

## II. DEEPFAKE GENERATION

Deepfake techniques involve several deep learning algorithm to generate fake contents in the form of videos, images, texts or voices. GAN are used to produce forged images and videos. Nowadays Deepfake technology become more popular due to its easiness and cheap availability. And also wide range of applications of Deepfake techniques create special interest among professional as well as novice users. A widely used model in deep network is deep autoencoders that has 2 uniform deep belief networks where 4 or 5 layers represent the encoding half and rest represent the decoding half. Deep encoding widely used in reduction of dimensions and compression of images [40] [41].First ever success of Deepfake technique was creation of FakeApp which is used to swap a person face with another person [42]. "FakeApp" software needs huge chunks of data for better results and data is given to the system to train the model and then source person face inserts into the targeted video. Creation of fake videos in FakeApp involves extraction of all images from source video in a folder, properly crop them and make alignment and then processed it with trained model, after merging the faces, the final video become ready [43]. Garrido et al. [44] proposed a method that use customized join structure method of the actor in the video as a shape representation, the overlay of blend shape model is refined in a temporally coherent way and after that high frequency shape detail reconstructed using photometric stereo approach that exceeds unknown light. Comparing with the technique discussed by Valgaerts et al. [45] that used binocular facial subject this method [44] show high shape details. Chen Cao et al. [46] proposed a process for practical facial finding and animation where video stream of source sent through user specific 3D shape regresser for 3D facial shape and for simultaneously compute face shape and motions, user blend shapes and camera matrix using the technique displaced dynamic expression(DDE) Chen Cao et al. [47] proposed a real time high fidelity facial capture system that able to reconstruct person specific ankle details in real time from a single camera. It automatically initializes person specific wrinkle probability map in any uncontrolled set of expressions.

It is based on global face tracker which generates low resolution mesh. Then the mesh tracking is improved and enhanced the details through novel local wrinkle regression methods. It can generates realistic facial wrinkles and also a globally plausible face shape which can be seen from any dimension. Comparing to Chen Cao et al. [46] it has low tracking error, with this error there are temporal noise and artifacts on final results. Comparing to high resolution capture method of Beeler et al. [48] this experiment did not capture every sort of details but this real time results is a plausible review without any depth constraints. Comparing to offline monocular capture method proposed by Garrido et al. [44] this experiment's results is clearly better in quality. Justus Thies et al. [49] introduced a method uses a new model based tracking approach based on a parametric face model. This model tracked from real time RGB-D input. In this experiment face model calibration for each actor needs before the tracking start. Since mouth changes shape in the target this experiment showed synthesizes new mouth interior using teeth proxy and texture of the mouth interior. Transfer quality is high even if the lighting quality in source and target differs in this method. Comparing with "FaceShift" of Weise et al. [50], although the geometric alignment of both approaches is similar, but this method achieve significantly better photo metric alignment. Comparing with Chen Cao et al. [46] (RGB only), this model's RGB-D approach reconstruct model that can shape expression details of the actor more closely.

Face2face, a real time facial reenactment method proposed by Justus Thies et al. [51] that works for any commodity webcam. Since this method uses only RGB data for source and target actor, it can manipulate real time Youtube video. A significant difference to previous method Justus Thies et al. [49] is the re rendering the mouth interior, for this, the authors designed internal side of lip of the target using video footage from the training sequence based on temporal and photo metric similarity. The system reconstruct and tracks both source and target actor using dense photo-metric energy minimization. Using a novel subspace deformation transfer technique it shifts appearance from origin to destination, then it re rendered the modified face on the top of target sequence in order to replace the original facial expressions. This method introduced new RGB pipeline which pre compared against modern real time face tracking process. Comparing to Thies el al. [49] and Cao et al. [46], it is noted that first one used RGB-D and second one and this method only use RGB data. In comparison with offline tracking algorithm of Shi et al. [52], it is found that that method used additional geometric refinement using shading cures. Comparing this method with Garrideo et al. [53] offline works, it is found that both method results similar reenactment data however this uses a geometric teeth proxy which leads artificially shift mouth region. Supasorn Suwajanakorn et al. [54] used the recurrent neural network that transform input video to time varying mouth shape. After synthesizing mouth texture the chips was enhanced in details.

Finally they blend the mouth texture on to a re time target video and match the pose. As this approach requires only audios it can generate high quality head close up even the original video has low resolution. This approach works on casual speech also. Comparing with the face2face [51], this method works better in some case such as face2face need input video to drive the animation of target person where this approach needs only audio speech. Justus Thies et al. [55] proposed a method that Enables full control over portrait video. It is based on short RGB-D video sequence of the destination person. From this sequence proxy mesh is extracted using voxel hashing. To generate an automated rig face template automatically fix to this mesh. Using dense correspondences between face mesh and scan mesh blend shaped has been transferred to proxy. And the automated rig enables fully animate the target proxy mesh. View dependent image synthesis proposed for deformed target actor .Then the target proxy mesh partitioned into three parts head, neck and body. For this three part the nearest texture retrieves from the initial video sequence. Then run the mesh into the space of target image compensating for the incomplete reconstruction by using image wrapped field. Then all three parts are composited to generate desire output image. Source actor expression and movement are tracked by using multi linear model for face and rigid proxy model for upper body and then the parameters are transferred to target actor's rig and rendered it. The eye texture are retrieved from the calibration sequence of target actor using the eye motion of source actor. Tab. 1 describes several Deepfake generation related article.

### A. Several Tools for Deepfake Generation

There are several tools and software available for Deepfake video generation. FaceSwap [56] is one of the open source software that use deep learning mechanism to swap faces in images or videos. Concept of Generative adversarial networks (GANS) is used which includes two encoder and decoder pairs. In this techniques parameters of encoders are shared. An improved version of previous tool is FaceSwap-GAN [57] ,that VGGface to Deepfakes auto-encoding architecture. It supports several output resolutions such as 64x64, 128x128, and 256x256. It is capable of realistic and consistent eye movements. It used VGGFace perceptual loss [58] concepts that helps to improve direction of eyeballs to be more precise and in line with input face. This tool includes MTCNN face detection technique[59] and kalman filter [59] to detect more stable face and smoothen it respectively. "Faceswap-pytorch" [60] another Deep- fake creation tool that makes dataset loader more efficient can load from 2 directory simultaneously. New face outline replace technique is used to get a much more combination result with original image. Most used tool for making Deepfake videos is "DeepFaceLab" [61], that runs much faster than previous version. It has more interactive converter. "DFaker" [62] is another tool that used DSSIM loss function [63] to reconstruct face.

TABLE I
SUMMARY OF REMARKABLE DEEPFAKE RELATED ARTICLES

| Serial | Paper Name | Authors | Date and Publication | Key findings | Limitations |
|---|---|---|---|---|---|
| 1 | Reconstructing detailed dynamic face geometry from monocular video[44] | Pablo Garrido, Levi Valgaert, Chenglei Wu, Christian Theobalt | ACM Trans. Graph. 32, 6, Article 158 (November 2013), 10 pages | Monocular tracking of facial expressions, In some cases, less susceptible to occlusions and drift, Well suited for video augmentation | Fail in extreme head angle, Not free from artifacts like tracking inaccuracies of teeth and lips |
| 2 | Displaced dynamic expression regression for real-time facial tracking and animation[46] | C. Cao, Q. Hou, and K. Zhou | ACM TOG, 33(4):43, 2014 | Robustness against fast motions, Large head rotations, lighting. Use of single web camera. | Not ideal for low quality images. |
| 3 | Real-Time High-Fidelity Facial Performance Capture[47] | Cao, C., Bradley, D., Zhou, K., Beeler | ACM Trans. Graph. 34, 4, Article 46 (August 2015) | Able to reconstruct person specific ankle details in real time from a single camera, Mesh tracking is improved and enhanced the details through novel local wrinkle regression methods. | Changing illumination can affect both optical flow and the local regresser, Problem of occlusions. |
| 4 | Real-time Expression Transfer for Facial Reenactment[49] | Thies, J., Zollhöfer, M., Nießner, M., Valgaerts, L., Stamminger, M., Theobalt, C. 2015 | ACM Trans. Graph. 34, 6, Article 183 (November 2015), 14 pages | Real time facial reenactment, This model tracked from real time rgb-d input, GPU based tracking | Tracking could fail if hands occludes the face, if occlusion is extreme the tracker will fail. |
| 5 | Face2Face: Real-time Face Capture and Reenactment of RGB Videos[51] | Justus Thies, Michael Zollhöfer, Christian Theobalt, Marc Stamminger, and Matthias Nießner | Proc. Computer Vision and Pattern Recognition (CVPR), IEEE 1 (2016) | RGB data input, Real time video manipulation, Using the temporal and photo-metric similarity | Hard shadows or Spectacular highlights, Capturing subjects with long hair and beard is very challenging, Mouth behavior can not learn if subject is too static or insufficient expressions |
| 6 | Synthesizing Obama: Learning Lip Sync from Audio[54] | Supasorn Suwajanakorn, Steven M. Seitz, and Ira Kemelmacher-Shlizerman. | ACM Trans. Graph. 36, 4, Article 95 (July 2017), 13 pages | Audio input is converted to a time-varying sparse mouth shape, Can synthesize visual speech only from audio. | In final composition, mouth shape and chin location can vary depending on target frame, Target person's non frontal face can cause mouth texture composited outside face, Face texture synthesis depends on complete mouth expressions. |

Most of the face swapping tools use the concepts of Generative adversarial networks(GANS). From [64, fig. 1 ], it is the visual diagram of encoder in GANS. It has the ability to extract deep information from image, in encoding stage it extracts the facial expressions of two individual. Then it analyze the common features between two faces and memorizes it. After that, in decoding stage in [64,fig.2], it decodes the information of two individuals. And in [64,fig. 3] shows the workings of discriminator that used to make decision of authenticity of features of every given data. Copy the source features to target features is most tedious task in face swapping and it is done by autoencoder after proper training. A implicit layer inside the encoder trained to generate the code to represent input. And auto encoder comprise of 2 slice encoder that represent input and decoder that generates the reconstruction. As represent in [64,fig. 4] and face swapping tools described here mostly used autoencoder technique. The main objectives of autoencoder is to handle MAE loss(reconstruction loss), adversarial loss( performs by Discriminator) and perceptual loss(which optimize similarity between source image and target image).

## III. DEEPFAKE DETECTION

It is often difficult sometimes impossible to detect Deepfake contents by a human being with a untrained eye. A good level of expertise is needed to detect irregularities in Deep fake videos. Till now there are several approaches have been proposed including machine detection, forensics, authentication as well as regulation to combat Deepfake.

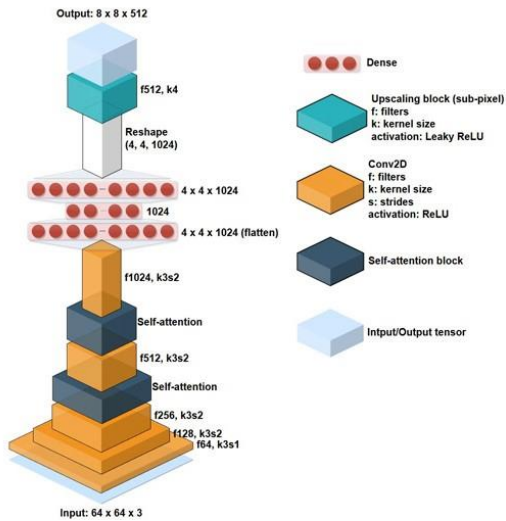

Fig. 1. Describe the encoding technique of face swap tools

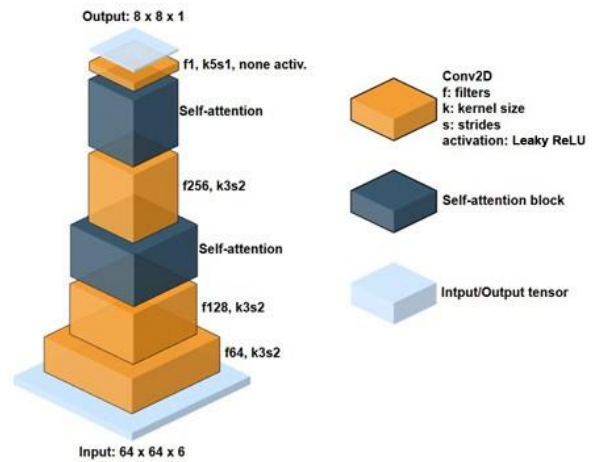

Fig. 3. Examine process of instances of given data

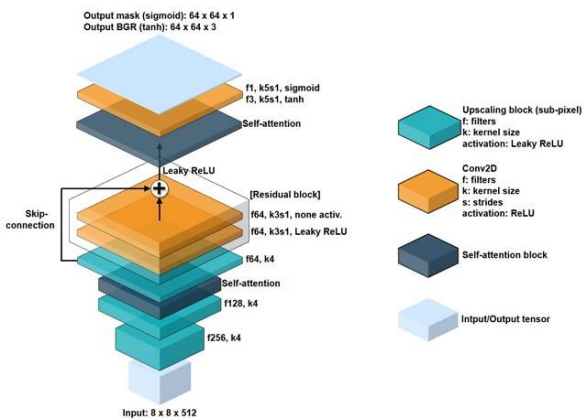

Fig. 2. Describe the decoding technique of face swap tools.

Experts say as Deepfake video is created by algorithm, while real video is made by a actual camera, it is possible to detect Deepfake from existing clues and artifacts. There are also some anomalies like lighting inconsistencies, image warping, smoothness in areas and unusual pixel formations which could help to detect Deepfake. Detecting Deepfakes Korshunov et al. [65] described a process that use to find inconsistency in the middle of visible mouth motion and voice in recording. In this article they also applied several approaches including simple principal component analysis (PCA) , linear discriminant analysis (LDA), image quality metrics (IQM) and support vector machine (SVM) [66]. VidTIMIT database [67], is a publicly available database of videos, used for generating several Deepfakes videos with combination of different features of Deepfake creation technique including face swapping, mouth movements, eye blinking etc. After using these videos to verify several methods of Deepfake detection, it is found that several face swap identification technique fail to detect fake contents.

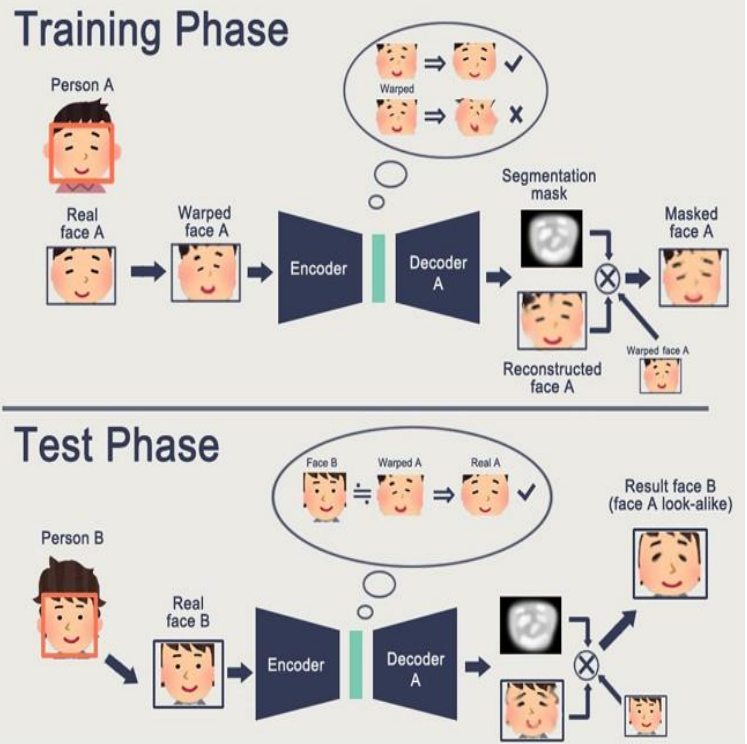

Fig. 4. Working procedures of autoencoder.

As example deep learning based face recognition technique VGG [68] and Facenet [69] are unable to detect Deepfakes properly. Although earlier Deepfake detection method was not able to measure the blink of eye, recent methods showed promising results to detect eye blink in source video and target video. The authors in [70] aimed to detect facial forgeries automatically and properly by using recent deep learning techniques convolutional neural networks (CNNs) with the help of neural network.

Forensics model face extreme difficulties to detect forgeries when the source data are made by CNN and GAN deep learning method [71]. To avoid the problems of adaptability, Huy H. Nguyen et al. proposed a method that supports generalization and could locate manipulated location easily. Ekraam Sabir et al. [72] tried to detect face swapping that generate by several available software such as Deepfake, face2face, faceswap by using conventional networks and recurrent unit. Deepfake contents could be in image format or video format. There are lots of research to detect fake videos and images. [73-76, Tab. 2] describes the several remarkable Deepfake articles that worked on detection.

### A. Detection based on several artifacts

There are lots of research that have done and thousands are ongoing to combat fake contents detection. Several artifacts such as head movements, facial expression, eye blinking are the most demanding subjects to the researcher in detecting fake videos. Yuezun Li et al. [73] have proposed a method that detect Deepfake using convolutional neural networks. In this paper they discussed the method to detect the distinction between fake video and real video based on resolution. In comparison with Headpose [74], this method shows better results as only detecting head pose does not work for frontal face. Li et al. [75] described a process that traced out eye blinking in fake video and detect the Deepfake videos based on CNN/RNN model. Shruti Agarwal et al. [76] discussed a forensic technique to detect Deepfakes by capturing facial expression and head movements. It showed better results from some other detection technique in some cases. In [73, fig. 5], it shows the detection of artifacts between original faces and generated faces by making the comparison between the face surroundings and other face regions with a dedicated CNN model.

### B. Binary Classifier based fake video detection

Huy H. Nguyen et al. [77] proposed capsule network based system to detect forged videos and images using the concepts of Capsule networks. Capsule networks able to generate the spatial relationships between several face parts and a complete face and by using concepts of deep convolutional neural networks, it able to detect various kinds of spoofs. In this paper the authors discussed the Replay Attack Detection, Face Swapping Detection, Facial Reenactment Detection pro cesses. Andreas Rössler et al. [78] discussed a detection tool that helps to find hoax contents generate by several deepfake generation tools such as Deepfakes, Face2Face, FaceSwap and Neural Texture. This method is able to check the region of face landmark perfectly. David Guera et al. [79] discussed the method of fake detection by using recurrent neural network where where they used CNN for features extraction on frame level and RNN to grab anomalies of frames. Mengnan Du et al. [80] proposed e Locality-aware Auto Encoder (LAE), that merge close depiction studying and implementing setting in a united framework. It makes predictions relying on correct evidence in order to boost generalization accuracy. Davide Cozzolino et al. [81] discussed a forgery detection technique that introduced a method based on neural network which is used to transfer between several manipulation domain. It also described how transferability enables robust forgery detection. In [ 78, fig.6 ], it shows the whole working process of previously discussed detection tool FaceForensic++ which used CNN model for reenactment, replacement and manipulation of fake contents.

Temporal discrepancies in the frames are discovered using R-CNN which combines convolutional network DenseNet [82] and recurrent unit cells [83]. LRCN based eye blinking [75], that learn temporal patterns of eye blinking, found frequency of blinking in Deepfake video is less than original. To discover resolution inconsistency between real face and its surrounding, face warping artifacts [73] is used that built on using techniques VGG 16 [84], ResNet50, 101 or 152 [85]. Capsule forensics [77] is introduced to extract features from face by VGG-19 network are classified by feeding into capsule networks. Through a numbers of looping, output of the three convolutional capsules is routed with the help of a dynamic routing algorithm [86] and as a result two output capsules has been generated, one for fake and another for real image. Combination of CNN and LSTM is applied in intra-frame and temporal inconsistencies [79] where CNN extracts fame level features and LSTM construct sequence descriptor. Pairwise learning [87] used CNN concatenated to CFFN Two-phase procedure one is feature extraction using Siamese network architecture [88] another is classification using CNN. MesoNet [89], which used the CNN technique ,introduced 2 deep networks Meso-4 and MesoInception-4 to detect fake video content. Generalization of GAN technique [90] used DCGAN,WGANGP and PGGAN enhances this generalization ability of model. It is used to remove low level features of images and focus on pixel to pixel comparison in the middle of hoax and actual images. Zhang et al. [91] introduced a classifiers that used SVM,RF,MLP to extract discriminant features and classify real vs fabricated. Eye, teach and facial texture [92] used logistic regression and neural network for classify several facial texture and compare with real one. PRNU Analysis [93] proposed to detect PRNU patterns between fake and authentic videos.

TABLE II
SUMMARY OF SEVERAL DEEPFAKE DETECTION RELATED ARTICLES

| Serial | Paper Name | Authors | Date and Publication | Key findings | Limitations |
|---|---|---|---|---|---|
| 1 | Exposing DeepFake Videos By Detecting Face Warping Artifacts[73] | Yuezun Li, Siwei Lyu | IEEE Xplore November, 2018 | Worked on visualize artifacts because of resolution problem between warped regions and surroundings, Used dedicated CNN method to trace out real face area and artifacts on warped face | Limited to face warped artifacts only, No direction for other anomalies like head moves or face occlusions. |
| 2 | Exposing Deep Fakes Using Inconsistent Head Poses[74] | Xin Yang, Yuezun Li and Siwei Lyu | In ICASSP, 2019 | Used SVM based classifer to accumulate feature vectors of targeted head | Not ideal for high resolution image. |
| 3 | In Ictu Oculi: Exposing AI Generated Fake Face Videos by Detecting Eye Blinking[75] | Yuezun Li, Ming-Ching Chang and Siwei Lyu | In IEEE International Workshop on Information Forensics and Security (WIFS), 2018 | Deploy the concepts of Long-term Recurrent Convolutional Networks(LRCN) to capture eye blinking. | No direction for fast eye blinking, eye blinking is the only way described here. |
| 4 | Protecting World Leaders Against Deep Fakes[76] | Shruti Agarwal and Hany Farid | In Proceedings of the IEEE Conference on Computer Vision and Pattern Recognition Workshops, 2019 | SVM based detection technique, analyzed all aspects of artifacts, used correlations between facial expressions and head moves. | Not work accurately in case of formal pose videos such as less head moves, minimize eye blinking. |

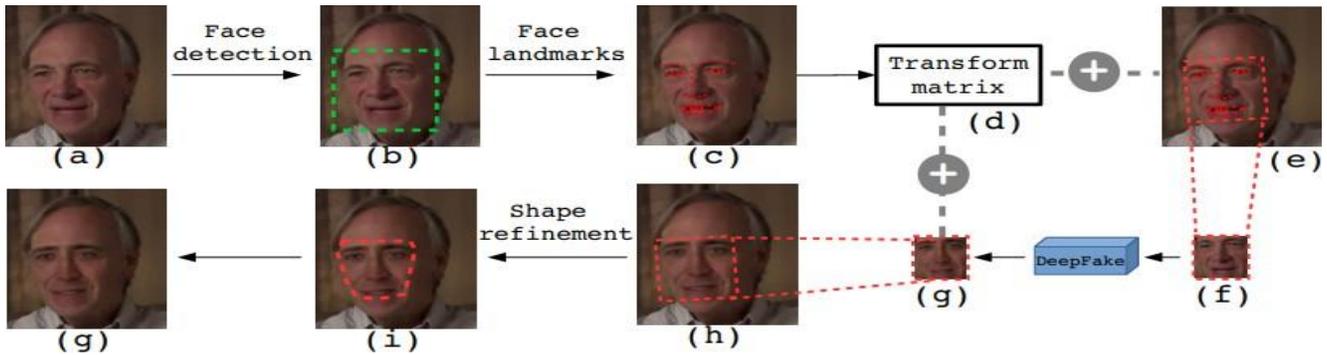

Fig. 5. Detecting Face Warping Artifacts.

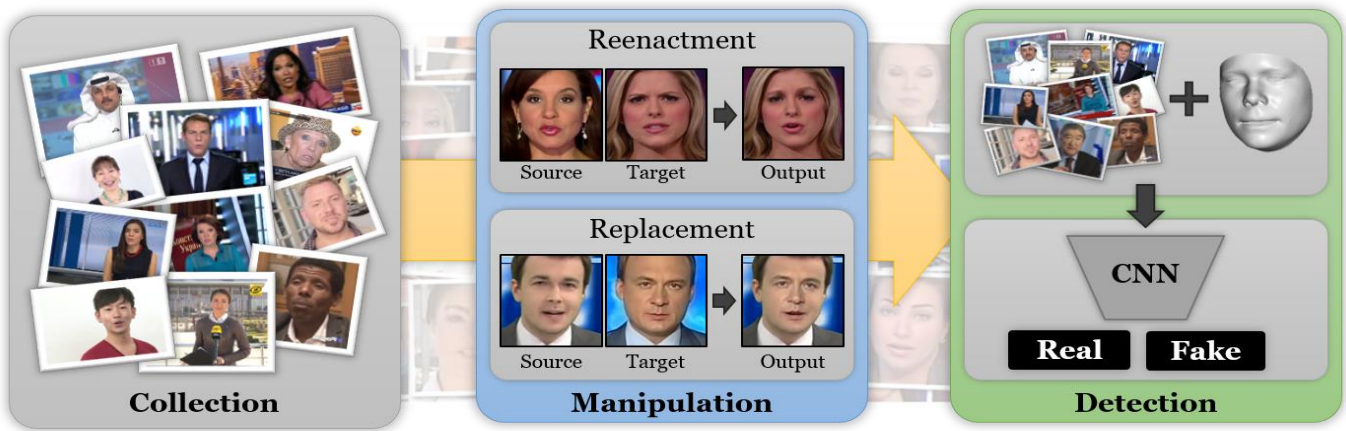

Fig. 6. Detection of manipulated image.

## IV. DISCUSSIONS AND CONCLUSIONS

In this paper, we have discussed several existing Deepfake videos and images generation and detection technique and application of Deepfakes both positive and negative. We also discussed several tools available for Deepfake creation and detection. Technology is making tremendous improvement day by day and every day new dimension of technology is coming to us.Improvement of Deepfake generation technique make the detection work difficult day by day. Improved detection techniques and accurate dataset are important issues for detecting Deepfake properly [94] [95]. From our study it is shown that Because of Deepfake technology, people are loosing trust on online content day by day. As Deepfake contents creation technique improving day by day, any person having a high configuration computer instead of having less technological knowledge could create Deepfakes content of any individuals for malicious purposes. Moreover, the advancement of internet and networking made it possible to spread Deepfake videos in a moment. This technique could influence the decisions of world leaders, important public figures which could be harmful for world peace [96]. From our study, we found that the battle between Deepfake creator and detector is growing rapidly. Although having lots of negativity, we have also discussed several positive use of Deepfake technology such as in film industry, fundraising etc. It is possible to return back the voice of a individual who have already lost it by using the Deepfake generation methods. There are lots of debate against and in favor of Deepfake technology. Our study tried to analyze the both sides of Deepfake technology. DeePfake technology have a detrimental effects on film industry, commercial media platform, social media. Deepfake technology used to gain trust of people via social media in any political context or any other social media context. To make profit by generating traffic of fake news via web platform is increasing rapidly [97]. Zannettou et al. [98] showed a number of people behind Deepfake including politicians, public figures, celebrity ,creating Deepfakes and spreading it via social media for various beneficial purposes of their own. We have also found that many organizations are involved in making Deepfake contents for useful purposes. According to our study Deepfake has lots of threats towards individuals, society, politics as well as business. Because of increasing fake contents , the situation become worse for media person to detect real one specially journalists.

On the other hand, we have found that proper development of anti development technology, proper rules regulation and awareness could combat Deepfake for malicious uses. Quick responses to fake contents uploaded in any social media platform is important to prevent further spreading. By building awareness among people and educate people about the literacy of Deepfake could lessen the expansion of malicious uses of this technology. Governments and several companies should run campaigning to build awareness against misuse of this technology. Furthermore, technological tools for detection and prevention Deepfake must be improved.

At the end, there are obviously some limitations of our findings. We have discussed certain articles, tools related to Deepfake technology. Through our paper we have reviewed quite good numbers of articles to understand the Deepfake generation techniques and current situation of this technology, its benefits and harms but still there are lot of scholarly papers ,articles of this technology to find more depth knowledge of this technology. Also we have just studied the articles briefly and tried to present the important aspects but extensive study would have present more better analysis. There are lots of analysis available of this technology and deep insights to this will provide better opportunities for further research in this field.